\documentclass{article}
\usepackage[utf8]{inputenc} % Compile UTF-8 file
\usepackage[margin=1in]{geometry} % Change paper settings
\usepackage{float} % Placing figures HERE
\usepackage{amsmath} % Math environments
\usepackage{amssymb} % Math logic symbols
\usepackage{bm} % \bold for greek letters
\usepackage{graphicx} % \includegraphics
\usepackage{hyperref} % hyperlinks for references
\usepackage{cleveref} % \cref for references
\usepackage{booktabs} % horizontal lines for tables
\usepackage{subcaption} % for images in a grid
\usepackage{tikz} % for Tikz diagrams
\usepackage{xcolor} % text colors
\usepackage{fancyhdr} % Custom headers and footers
\usepackage{mathtools} % \mathclap and \substack for long summation parameters
\usepackage{algorithm}
\usepackage[noend]{algpseudocode}

\def\bm{\boldsymbol}

\newcommand{\comment}[1]{}

\newcommand{\BEA}{\begin{eqnarray}}
\newcommand{\EEA}{\end{eqnarray}}

\newcommand{\mbp}{\mathbf{p}}
\newcommand{\mbq}{\mathbf{q}}

\newcommand{\BR}{\mathbb{R}}

\newcommand{\BH}{\mathcal{H}}
\newcommand{\Be}{\mathbf{e}}

\newcommand{\di}{\mathrm{d}}

\newcommand{\BBE}{\mathop{\mathbb{E}}}
\newcommand{\tree}{\mathcal{T}}

\newcommand{\Btheta}{\bm{\theta}}
\newcommand{\BTheta}{\bm{\Theta}}

\newcommand{\Integrator}{\mathrm{Integrator}}
\newcommand{\energychange}{E_{\text{rel}}}
\newcommand{\binaryset}{\mathbb{B}}
\newcommand{\unaryset}{\mathbb{U}}
\newcommand{\interactionset}{\mathbb{I}}
\newcommand{\norm}[1]{\left\lVert#1\right\rVert}

\date{}
\title{H-FEX: A Symbolic Learning Method for Hamiltonian Systems}

\author{Jasen Lai\\
    Department of Mathematics, University of Florida,\\
    1400 Stadium Rd, Gainesville, FL 32611, USA\\
    {\tt lai.jasen@ufl.edu }  \vspace{0.1in} \\
    Senwei Liang\\
    Lawrence Berkeley National Laboratory, Berkeley, CA 94720\\
    {\tt SenweiLiang@lbl.gov}
    \vspace{0.1in} \\
    Chunmei Wang\\
    Department of Mathematics, University of Florida,\\
    1400 Stadium Rd, Gainesville, FL 32611, USA\\
    {\tt chunmei.wang@ufl.edu} 
} 
\begin{document}
\maketitle

\begin{abstract}

Hamiltonian systems describe a broad class of dynamical systems governed by Hamiltonian functions, which encode the total energy and dictate the evolution of the system. Data-driven approaches, such as symbolic regression and neural network-based methods, provide a means to learn the governing equations of dynamical systems directly from observational data of Hamiltonian systems. However, these methods often struggle to accurately capture complex Hamiltonian functions while preserving energy conservation. To overcome this limitation, we propose the Finite Expression Method for learning Hamiltonian Systems (H-FEX), a symbolic learning method that introduces novel interaction nodes designed to capture intricate interaction terms effectively. 
Our experiments, including those on highly stiff dynamical systems, demonstrate that H-FEX can recover Hamiltonian functions of complex systems that accurately capture system dynamics and preserve energy over long time horizons.
These findings highlight the potential of H-FEX as a powerful framework for discovering closed-form expressions of complex dynamical systems.

\end{abstract}

\section{Introduction}
Symbolic regression is a technique for discovering mathematical equations from data, 
making it a useful tool for recovering the underlying laws of physical systems \cite{SINDY}.
Unlike traditional regression, which fits parameters to a predefined functional form, 
symbolic regression searches the space of mathematical expressions to find the model that best fits the dataset.
One important class of dynamical systems where such methods are highly applicable is Hamiltonian systems \cite{NNSymplecticIntegrationSymbolicLearningHamiltonianSystems-1, NNSymplecticIntegrationSymbolicLearningHamiltonianSystems-2}, 
which describe a wide range of physical phenomena, 
including celestial mechanics \cite{NBodyProblem}, quantum mechanics \cite{Hamiltonian2QuantumMechanics}, and control systems \cite{HamiltonianForControlSystems}.
These systems are governed by Hamiltonian functions,
which encode the total energy of the system 
and determine its evolution through Hamilton’s equations.
A key aspect of modeling Hamiltonian systems is energy conservation \cite{EnergyConservation}, 
which is essential for ensuring physically meaningful behavior, especially over long time horizons.
Given their importance, 
Hamiltonian systems are a natural setting for symbolic regression methods,
which can discover Hamiltonian functions directly from data.

While there is a need for data-driven discovery of Hamiltonian functions, existing methods struggle to balance interpretability with adherence to energy conservation laws.
A well-known approach for discovering governing equations from data is sparse-identification of nonlinear dynamics (SINDy) \cite{SINDY}, which recovers equations by selecting terms from a predefined set of basis functions.
However, on its own, SINDy does not enforce Hamiltonian dynamics,
and its reliance on linear combinations of basis functions limits its ability to represent complex Hamiltonian functions.
In contrast, Hamiltonian Neural Networks (HNNs) \cite{HNN} explicitly enforce Hamiltonian dynamics by using a neural network surrogate of the Hamiltonian function, ensuring energy conservation. 
Despite this advantage, HNNs lack interpretability, as their learned representations are encoded in black-box neural networks.
More advanced variants, such as Symplectic Recurrent Neural Networks (SRNNs) \cite{SRNN}, which incorporate multi-step integration, and Stiffness-Aware Neural Networks (SANNs) \cite{SANN}, which can identify stiff portions of training data, improve robustness and accuracy. However, they still rely on neural networks, which makes it difficult to produce meaningful closed-form expressions and to ensure accurate long-term predictions. 
These limitations highlight the need for a symbolic regression method tailored for Hamiltonian systems, one that can learn complex Hamiltonian functions accurately while respecting conservation laws.

Finite Expression Method (FEX) is a symbolic regression method that leverages reinforcement learning to discover complex closed-form expressions from data \cite{FEX-Jiang, FEX-Liang, FEX-Song} using reinforcement learning. 
Unlike SINDy, which relies on linear combinations of predefined basis functions, FEX represents expressions as trees of operators with associated weights. This structure enables it to capture a broader class of mathematical expressions, including those involving function composition. 
However, FEX is not designed to explicitly enforce Hamiltonian dynamics, making a direct application to Hamiltonian systems challenging. 
Additionally, many Hamiltonian systems \cite{ThreeBodyProblem, CoupledHarmonicOscillators,Hamiltonianexp} contain interaction terms that represent dependencies between multiple interacting objects.
FEX lacks a way to model these interaction terms, further limiting its ability to model complex Hamiltonian systems.

To address the shortcomings, we introduce Finite Expression Method for Hamiltonian Systems (H-FEX), an adaptation of FEX specifically designed for Hamiltonian systems.
H-FEX modifies the search loop of FEX to enforce Hamiltoniannamics, similar to how HNNs impose Hamiltonian constraints to respect conservation laws.
Moreover, we introduce interaction nodes, 
which enable the construction of interaction terms 
that capture dependencies between interacting objects.
Through numerical experiments, we demonstrate that H-FEX accurately recovers interpretable, closed-form representations of the Hamiltonian function.
Additionally, H-FEX outperforms other methods by generating trajectories
that are highly accurate and preserve energy over long time horizons.
By bridging the gap between FEX and Hamiltonian systems, 
H-FEX is a powerful tool for discovering closed-form representations of Hamiltonian systems directly from data.

The remainder of this paper is organized as follows.
\Cref{sec:problem_statement} introduces Hamiltonian systems and defines the loss function used to train a Hamiltonian surrogate.
\Cref{sec:methodology} provides an overview of H-FEX, describes the interaction nodes, and explains how integrators are used during training and evaluation.
\Cref{sec:numerical_results} presents experiments benchmarking H-FEX against existing methods.
Finally, \Cref{sec:conclusion} summarizes the contributions and outlines directions for future research.

\section{Problem Statement} \label{sec:problem_statement}
A Hamiltonian system is a dynamical system
characterized by the Hamiltonian function $\BH(\mbp, \mbq)$, which is a real-valued function $\BH: \BR^{2d} \to \BR$ that maps the momentum $\mbp\in \BR^d$
and the position $\mbq \in \BR^d$
to a scalar energy.
A Hamiltonian system is said to be separable if it can be written as $\BH(\mbp, \mbq)=K(\mbp) + U(\mbq)$, where $K$ and $U$ denote the kinectic and potential energy, respectively. 
The phase space $\BR^{2d}$ represents the space of all possible states in the system.
The time evolution of the system is governed by Hamilton's equations:
\begin{align}
    \frac{\di \mbp }{\di t} = -\frac{\partial \BH}{\partial \mbq}, \quad 
    \frac{\di \mbq }{\di t} = \frac{\partial \BH}{\partial \mbp}.
    \label{eq:hamiltonian}
\end{align}
Trajectories in a Hamiltonian system describe how a particle evolves over time, starting from an initial condition.

Integrators are numerical methods that compute trajectories over a time interval, given an initial condition \cite{NumericalIntegrators}.
 Consider the discretization of an interval $[0, T]$
using uniformly spaced time points $(t_n)_{n=0}^N$, where $t_0=0$ and $t_N=T$.
Given a Hamiltonian function $\BH$ and an initial condition $(\mbp_{t_0}, \mbq_{t_0})$, 
an integrator computes the trajectory evaluated at time points $(t_n)_{n=0}^N$:
\begin{equation}
    (\mbp_{t_n}, \mbq_{t_n})_{n=0}^N=\Integrator(\BH, (\mbp_{t_0}, \mbq_{t_0}), (t_n)_{n=0}^N).
\end{equation}

Given observed trajectories $(\mbp_{t_n}, \mbq_{t_n})_{n=0}^N$ from an unknown Hamiltonian system,
where the initial condition $(\mbp_{t_0}, \mbq_{t_0})$ lies in a domain $ \Omega \subseteq \BR^{2d}$, 
the goal is to learn a parameterized surrogate Hamiltonian $\hat{\BH}_{\BTheta}$ that approximates the true, unknown Hamiltonian $\BH$, thereby recovering
the corresponding Hamiltonian dynamics. 
Using an integrator, predicted trajectories can be generated from the surrogate $\hat{\BH}_{\BTheta}$ starting from the initial condition $(\mbp_{t_0}, \mbq_{t_0})\in \Omega$:
\begin{equation}
    (\hat{\mbp}_{t_n}, \hat{\mbq}_{t_n})_{n=0}^N=\Integrator(\hat{\BH}_{\BTheta}, (\mbp_{t_0}, \mbq_{t_0}), (t_n)_{n=0}^N).
    \label{eq:predicted_trajectory}
\end{equation}
Ideally, the surrogate Hamiltonian function should produce predicted trajectories
that closely match the observed trajectories given the same initial conditions.
To achieve this, we learn $\hat{\BH}_{\BTheta}$ by minimizing the loss function:
\begin{equation}
    \mathcal{L}(\BTheta) = \BBE_{(\mbp_{t_0}, \mbq_{t_0}) \sim \mathbb{P}_{\Omega}} \left[ \sum_{n=0}^N \left[ \norm{\hat{\mbp}_{t_n}-\mbp_{t_n}}^2 + \norm{\hat{\mbq}_{t_n}-\mbq_{t_n}}^2 \right] \right],
    \label{eq:loss}
\end{equation}
where $(\hat{\mbp}_{t_n}, \hat{\mbq}_{t_n})_{n=0}^N$ are generated by $\hat{\BH}_{\BTheta}$ using \Cref{eq:predicted_trajectory}
and $\mathbb{P}_{\Omega}$ is some probability distribution over the domain $\Omega \subseteq \BR^{2d}$ from which the initial conditions are sampled.

In practise, we have a dataset $\mathcal{D}$ of multiple observed trajectories, where each trajectory $(\mbp_{t_n}, \mbq_{t_n})_{n=0}^N$ has an initial condition sampled from a domain $(\mbp_{t_0}, \mbq_{t_0}) \sim \mathbb{P}_{\Omega}$. 
We can learn a surrogate $\hat{\BH}_{\BTheta}$ by minimizing the empirical loss function: 
\begin{equation}
    \hat{\mathcal{L}}(\BTheta) = \frac{1}{N|\mathcal{D}|}\sum_{(\mbp_{t_n}, \mbq_{t_n})_{n=0}^N \in \mathcal{D}} \left[ \sum_{n=0}^N \left[ \norm{\hat{\mbp}_{t_n}-\mbp_{t_n}}^2 + \norm{\hat{\mbq}_{t_n}-\mbq_{t_n}}^2 \right] \right],
    \label{eq:empirical_loss}
\end{equation}
where $|\mathcal{D}|$ denotes the number of training trajectories and $(\hat{\mbp}_{t_n}, \hat{\mbq}_{t_n})$ are generated by $\hat{\BH}_{\BTheta}$ using \Cref{eq:predicted_trajectory}.  
By minimizing \Cref{eq:empirical_loss}, we learn a surrogate Hamiltonian function $\hat{\BH}_{\BTheta}$ which can be used to model an unknown Hamiltonian system.
 
 \section{Methodology} \label{sec:methodology}
 \subsection{Finite Expression Method for Hamiltonian Systems}
H-FEX is a  FEX     method \cite{FEX-Jiang, FEX-Liang, FEX-Song}  to build a symbolic approximation
using mathematical expression with a finite number of operators, 
or more simply called finite expressions.
A finite expression is a combination of operators 
(e.g., $\times$, $\exp$, $(\cdot)^2$, $(\cdot)^4$) 
that form a valid function (e.g., $f(p, q) = \exp(p^2\times q^4)$).
If we limit our operators to unary operators and binary operators, 
we can represent a finite expression with a binary tree $\tree$
consisting of unary and binary nodes. 
In summary, H-FEX is a method that models the Hamiltonian function using a closed-form expression.

Given a tree $\tree$, 
we can assign operators to the nodes by an operator sequence $\Be$, using inorder traversal. 
In FEX, each element of the operator sequence $\Be$ is either from a set of unary operators $\unaryset$ (e.g., $\unaryset = \{ \exp, \sin, \mathrm{Id}, (\cdot)^2 \}$) or from a set of binary operators $\binaryset$ (e.g., $\binaryset = \{ \times, +, \div \}$), where the operators are applied element-wise to vectors.
In addition, we parameterize the tree with weights $\Btheta$
to expand the class of functions that can be represented by FEX. 
These weights can be applied between nodes in the tree, influencing intermediate values at various points in the resulting expressions. This enables H-FEX to learn specific constants in the governing equations.
In summary, a FEX approximation of the Hamiltonian function $\hat{\BH}(\mbp, \mbq; \tree, \Be, \Btheta)$ 
is a function of $(\mbp, \mbq)$ with parameters $\tree,\Be,\Btheta$.

\begin{figure}
    \centering
    \includegraphics[width=\linewidth]{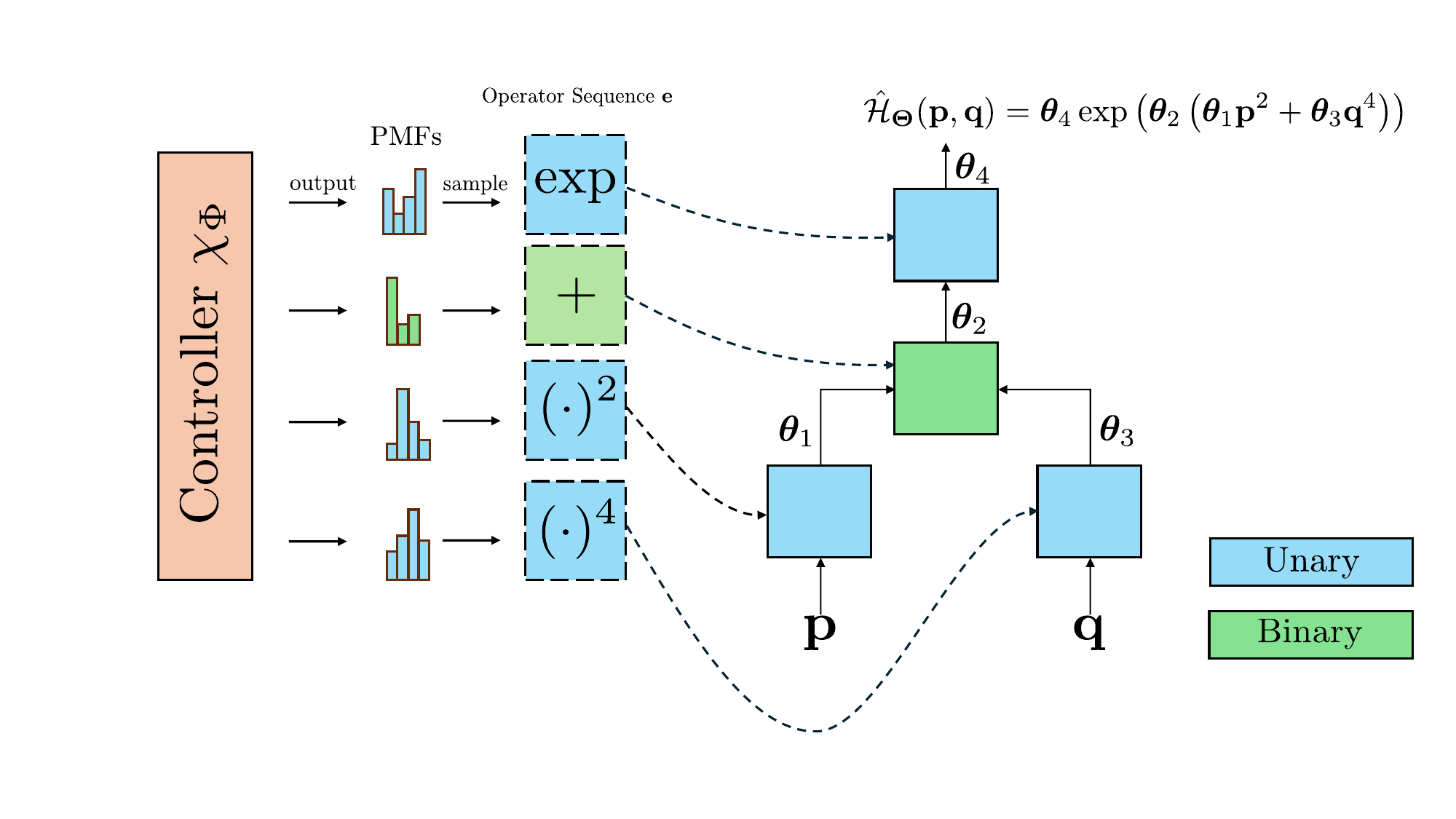}
    \caption{\textbf{Generating operator sequences for an H-FEX tree.} To parameterize an H-FEX tree with an operator sequence $\Be$, the controller $\chi_\Phi$ generates a PMF  for each node. An operator is then sampled from each PMF and assigned to its corresponding node. The resulting H-FEX tree becomes an expression with trainable weights, which are optimized during the scoring process in \Cref{eq:score_computation}. }
    \label{fig:fex-diagram}
\end{figure}

When H-FEX learns the surrogate Hamiltonian function $\hat{\mathcal{H}}_{\BTheta}$ in \Cref{eq:empirical_loss}, we let the trainable parameters be $\BTheta = (\Be, \Btheta)$, where the binary tree structure $\tree$ is fixed before training.
Minimizing \Cref{eq:empirical_loss} over both discrete and continuous parameters $\BTheta = (\Be, \Btheta)$ is a difficult task \cite{MixedOptimization}. 
The operator sequence $\Be$ consists of operators selected from some discrete space of $\unaryset$ and $\binaryset$, 
and the weights $\Btheta$ are real-valued weights from a continuous space.
To solve this mixed optimization problem, 
we use a search loop, based on reinforcement learning (RL),
to identify effective operator sequences $\Be$ while
optimizing weights $\Btheta$. 
The search loop is summarized in the following steps.

\begin{enumerate}
    \item \textbf{Operator sequence generation.}
    We use RL techniques to identify good candidate operator sequences $\Be$. 
    In RL, 
    a controller outputs a policy, 
    a probability mass function (PMF),
    and an action is sampled according to the probabilities \cite{ReinforcementLearning}.
    In our context, an action corresponds to an operator sequence $\Be=(e_1,\dots,e_Q)$, 
    where $Q$ denotes the number of operator nodes in tree $\tree$. 
    
    The controller $\chi_{\Phi}$ is a fully-connected neural network
    parameterized by weights $\Phi$
    and outputs PMFs $(p_1,\dots,p_Q)$. 
    For the $i$th node,
    we sample an operator from the corresponding PMF; i.e.,  $e_i \sim p_i$, 
    where $1\le i \le Q$,
    giving us a sampled operator sequence $\Be$. 
    As shorthand for the above process, 
    we write $\Be \sim \chi_\Phi$ to denote an operator sequence being sampled from the controller. 
    During a single iteration of the search loop,
    we sample $M$ operator sequences $\Be$ from the controller.
    
    To enhance the exploration of the operator sequence space, 
    we sample using an $\epsilon$-greedy strategy \cite{EpsilonGreedyStrategy}, 
    where $0 <\epsilon<1$. 
    With probability $\epsilon$, 
    we sample $e_i$ from a uniform distribution, 
    and with probability $1-\epsilon$, 
    we sample $e_i$ from PMF $p_i$. 
    A larger $\epsilon$ increases the likelihood of exploring new sequences.

    \item \textbf{Score computation.} \label{sec:score_computation}
    Given an operator sequence $\Be$,
    we compute a score $S(\Be)$ to quantify the performance of $\Be$:
    \begin{equation}
        S(\Be)\coloneqq\frac{1}{1+L(\Be)},
        \label{eq:score_computation}
    \end{equation}
    where $L(\Be)\coloneqq \min_{\Btheta} \hat{\mathcal{L}}((\Be, \Btheta))$ is the minimum loss optimizing over weights $\Btheta$ given a fixed operator sequence $\Be$, and $\hat{\mathcal{L}}$ is defined in \Cref{eq:empirical_loss}.
    As $L(\Be)$ approaches 0, 
    the score $S(\Be)$ approaches 1,
    meaning a larger score indicates a better performing operator sequence $\Be$. In practice, we approximate $L(\Be)$ with the empirical minimum loss attained when optimizing for a fixed number of steps.

    From the previous step, we sampled $M$ operator sequences from the controller: $\Be_1,\ldots, \Be_M \sim \chi_\Phi$, 
    and compute the corresponding scores $S(\Be_1),\ldots, S(\Be_M)$ for each sequence individually. 
    During score computation,
    we typically trade off accuracy for speed when approximating $L(\Be)$
    to expedite the exploration of operator sequences \cite{FEX-Song}.
    As a result, the weights $\Btheta$ learned during this process may not be optimal. 
    We address this issue in the next step.

    \item \textbf{Candidate Pool.}
    We maintain a candidate pool to keep track of the highest-scoring operator sequences.
    The candidate pool stores the parameters $\BTheta = (\Be, \Btheta)$, obtained after score computation, for the top $K$ scoring operator sequences $\Be$ across all iterations of the search loop. 
    Once the search loop concludes,
    we fine-tune the models in the candidate pool $\mathcal{K}$
    by further minimizing $\mathcal{L}((\Be, \Btheta))$
    with respect to $\Btheta$,
    if necessary. 

    \item \textbf{Controller update.}
    We use a risk-seeking policy gradient \cite{RiskSeekingPolicyGradient} to update the controller $\chi_\Phi$. 
    Standard RL techniques update the weights $\Phi$ of the controller
    $\chi_\Phi$ by maximizing the objective:
    \begin{equation}
        \mathcal{J}_s(\Phi)\coloneqq \underset{\Be \sim \chi_{\Phi}}{\mathbb{E}}[S(\Be)].
    \end{equation}
    This objective function corresponds to maximizing the average score of operator sequences sampled by the controller $\chi_\Phi$.
    However, in symbolic regression, our interest lies in the best-forming operator sequences rather than the average.
    Therefore, we instead maximize the expected score of the top $(1-\nu)\times100\%$ scores of operator sequences  sampled from the controller $\chi_\Phi$:
    \begin{equation} \label{eq:risk-seeking-objective}
        \mathcal{J}(\Phi)\coloneqq \underset{\Be \sim \chi_\Phi}{\mathbb{E}}[S(\Be) | S(\Be) \ge S_\nu(\Phi)],
    \end{equation}
    where $S_\nu(\Phi)$ is the $(1-\nu)\times 100\%$ quantile of the score distribution of operator sequences sampled from $\chi_\Phi$.

    We estimate the gradient of \Cref{eq:risk-seeking-objective} using the risk-seeking policy gradient in \cite{RiskSeekingPolicyGradient}:
    \begin{equation} \label{eq:risk-seeking-gradient}
        \nabla_\Phi\mathcal{J}(\Phi) \approx \frac{1}{\nu M}\sum_{j=1}^M [S(\Be^{(j)})-\tilde{S}_\nu(\Phi)] \cdot \mathbf{1}\{S(\Be^{(j)})\ge \tilde{S}_\nu(\Phi)\} \cdot  \nabla_\Phi \sum_{i=1}^Q \log p_i(e^{(j)}_i;\Phi),
    \end{equation}
    where $p_i(\cdot;\Phi)$ is the PMF of the $i$th node generated by the controller $\chi_\Phi$, and 
    $\tilde{S}_\nu(\Phi)$ is the empirical $(1-\nu)\times 100\%$ quantile
    of the $M$ sampled operator sequences $\Be^{(1)},\dots,\Be^{(M)} \sim \chi_\Phi$.
    At the end of each iteration of the search loop, we compute the gradient in \Cref{eq:risk-seeking-gradient} and update the weights $\Phi$ of the controller $\chi_\Phi$ accordingly. 
\end{enumerate}

\subsection{Interaction Nodes}
Hamiltonian systems involving multiple objects (e.g., particles, oscillators) typically have Hamiltonian functions that include interaction terms among position coordinates $\mbq_i$, and the functional forms of momentum $\mbp$ and position $\mbq$ often differ.
For example, in the N-body problem \cite{NBodyProblem}, the Hamiltonian function is given by $\BH(\mbp_1, \dots, \mbp_N, \mbq_1,\ldots, \mbq_N)=\sum_{i=1}^N c_0\lVert \mbp_i \rVert^2 - \sum_{i<j}\frac{c_1}{\lVert \mbq_i - \mbq_j \rVert}$,
and in coupled harmonic oscillators \cite{CoupledHarmonicOscillators}, it is $\BH(\mbp_1, \mbp_2, \mbq_1, \mbq_2)=c_0\mbp_1^2+c_1\mbp_2^2+c_2\mbq_1^2+c_3\mbq_2^2+c_4\mbq_1\mbq_2$, 
where $c_i$ are constants. 
Many Hamiltonian functions exhibit these structures, so we design the H-FEX tree to treat $\mbp$ and $\mbq$ separately and introduce a dedicated interaction node to effectively represent interaction terms. 

\begin{figure}
    \centering
    \includegraphics[width=0.9\linewidth]{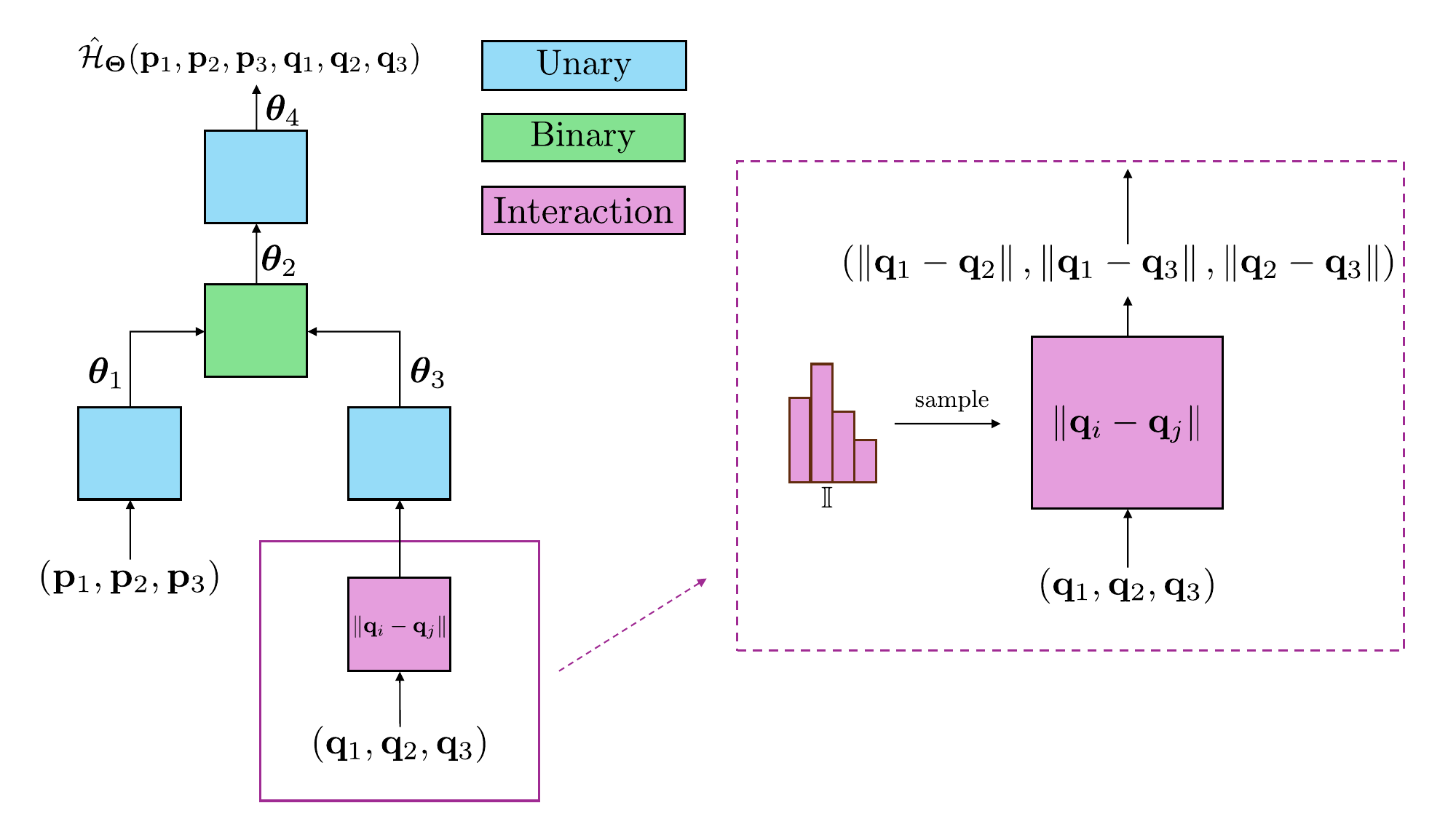}
    \caption{\textbf{H-FEX tree with an interaction node.} This example H-FEX tree includes an interaction node that operates on the position coordinates $(\mbq_1, \mbq_2, \mbq_3)$. The interaction node is parameterized by an operator sampled from a PMF over the set of interaction operators $\mathbb{I}$.
    The set $\mathbb{I}$ contains operators that maps pairwise combinations of elements to a scalar value.
    In this instance, the sampled interaction operator is $\norm{\mbq_i-\mbq_j}$ for $i \ne j$, and the node outputs all pairwise interactions between position coordinates.}
    % {\color{redf} what is dictionary $\mathbb{I}$?}
    \label{fig:interaction}
\end{figure}

The interaction node enables an H-FEX tree to represent pairwise interaction terms.
Similar to unary and binary nodes,
the interaction node is assigned an operator sampled from a set $\mathbb{I}$.
Each operator in $\mathbb{I}$ maps a pairwise combination of elements to a scalar value.
If the interaction node takes a vector input $\mathbf{x} \in \BR^d$, then the set $\mathbb{I}$ may include operators such as $(x_i-x_j)$, $(x_i-x_j)^2$, $x_ix_j$, where $x_i$ and $x_j$ are distinct elements of $\mathbf{x}$ (i.e., $i \neq j$).
More generally, if input is a matrix $\mathbf{X} \in \BR^{r \times d}$, then the set $\mathbb{I}$ may include operators such as $\Vert \mathbf{x}_i-\mathbf{x}_j \rVert$, $\Vert \mathbf{x}_i-\mathbf{x}_j \rVert^2$, $\lVert\mathbf{x}_i \odot \mathbf{x}_j\rVert$, where $\norm{\cdot}$ denotes the Euclidean norm and $\odot$ denotes elementwise multiplication. 
These interaction operators map each pairwise combination of the $r$ column vectors in $X$ to a scalar.
In essence, the interaction node enables the H-FEX tree to model interaction terms, allowing it to capture complex Hamiltonian functions that involve pairwise interactions. \Cref{fig:interaction} shows an example of an H-FEX tree with an interaction node to represent interaction terms among the position coordinates.
% {\color{redf}  explain $\odot$ and $\|\cdot\|$}

\subsection{Training and evaluation for H-FEX}
Integrators enable the simulation of trajectories of candidate H-FEX trees during both training and evaluation. 
Once an operator sequence $\Be$ parameterizes an H-FEX tree, the resulting expression yields a Hamiltonian surrogate $\BH_{\BTheta}$, which can then be integrated into the predicted trajectories generated. During training, these trajectories are compared to observed data to calculate the empirical loss in \Cref{eq:empirical_loss}, which is also used to calculate a score of the operator sequence $\Be$ in \Cref{eq:score_computation}.
During the evaluation, a trained Hamiltonian surrogate $\hat{\BH_{\BTheta}}$ is used to simulate the trajectories of a given initial condition. At a high level, the prediction of the trajectory is summarized in \Cref{eq:predicted_trajectory}, but in this section, we provide a more detailed explanation in the following. 

To reduce numerical error during integration, we adopt a multi-step integration scheme similar to that used in \cite{SRNN}. 
Rather than integrating from $t_{n-1}$ to $t_n$ in a single step, each interval $[t_{n-1}, t_n]$ is subdivided into uniform $K$ substeps $(t_{n_k})_{k=0}^K$. 
During evaluation, trajectories are recursively generated by integrating forward from the previously predicted state:
\begin{equation}
    (\hat{\mbp}_{t_{n+1}}, \hat{\mbq}_{t_{n+1}}) = \Integrator (\hat{\BH}_{\BTheta},(\hat{\mbp}_{t_{n}}, \hat{\mbq}_{t_{n}}), (t_{n_k})_{k=0}^K) \quad  \text{for} \; n=0,1,\ldots,N-1,
    \label{eq:eval_integrator}
\end{equation}
where $(\hat{\mbp}_{t_{0}}, \hat{\mbq}_{t_{0}})$ is set to the initial condition.
In contrast, training uses the previous state from the observed trajectory at each step:
\begin{equation}
    (\hat{\mbp}_{t_{n+1}}, \hat{\mbq}_{t_{n+1}}) = \Integrator (\hat{\BH}_{\BTheta},(\mbp_{t_{n}}, \mbq_{t_{n}}), (t_{n_k})_{k=0}^K) \quad  \text{for} \; n=0,1,\ldots,N-1,
    \label{eq:train_integrator}
\end{equation}
which prevents the propogation of integration errors during optimization, leading to more stable training.
In \Cref{eq:eval_integrator} and \Cref{eq:train_integrator}, we use the Leapfrog integrator in \Cref{alg:leapfrog} for separable systems and the second-order Runge-Kutta (RK2) method in \Cref{alg:rk2} for non-separable systems. 
Additional details on integrators are provided in \Cref{apx:integrator_details}.
% Implementation details for both integrators are provided in \Cref{sec:leapfrog} and \Cref{sec:rk2} respectively. 
% {\color{redf} add a reference for Leapfrog integrator here}

\section{Numerical results} \label{sec:numerical_results}
In this section, we evaluate the performance of H-FEX by comparing it with existing methods on two problems: a nonseparable Hamiltonian system \cite{NonseparableProblem} and the three-body problem \cite{ThreeBodyProblem}.
In both cases, the objective is to learn the underlying Hamiltonian function directly from trajectory data, thereby recovering the corresponding Hamiltonian system. 

The experiments are conducted as follows. 
We begin by generating trajectores from the true system
and use them as training data for each method.
After training,
the learned models are evaluated on a separate test set consisting of trajectories not seen during training.
Using the initial states of the test data,
we simulate predicted trajectories from the learned surrogate of each method
and compare them to the corresponding true test trajectories.
The test trajectories span significantly longer time intervals than those used for training, allowing us to assess each model's ability to generalize and remain stable over extended predictions.
We evaluate the methods based on the accuracy of the predicted trajectories over time and the degree to which they conserve energy. 

We assess the accuracy of a predicted trajectory $(\hat{\mbp}_{t_i}, \hat{\mbq}_{t_i})_{i=0}^{N_{\text{test}}}$ by calculating the mean squared error (MSE) at each time step:
\begin{equation}
    \mathrm{MSE}(t) \coloneqq \frac{1}{2d} \left( \left| \left|\hat{\mbp}_{t} - \mbp_{t} \
\right| \right|^2 + \left| \left|\hat{\mbq}_{t} - \mbq_{t} \right| \right|^2 \right),
\label{eq:MSE-time}
\end{equation}
where $\mbp_t,\hat{\mbp}_t,\mbq_t, \hat{\mbq}_t \in \BR^d$. 
For each test trajectory, we compute $\mathrm{MSE}(t)$ at each time point to analyze how the prediction error evolves over time. Additionally, we assess the ability of each method to conserve energy by computing the relative error, denoted by $\energychange(t)$ in the energy over time: 
\begin{equation}
\energychange(t) \coloneqq \frac{|\BH(\hat{\mbp}_{t}, \hat{\mbq}_{t}) - \BH(\hat{\mbp}_{t_0}, \hat{\mbq}_{t_0})|}{|\BH(\hat{\mbp}_{t_0}, \hat{\mbq}_{t_0})|},
\label{eq:relerr-time}
\end{equation}
where $|\cdot|$ denotes absolute value and $\BH$ is the true Hamiltonian function and $(\hat{\mbp}_{t_0}, \hat{\mbq}_{t_0})$ is the initial state.
If the predictions $(\hat{\mbp}_{t}, \hat{\mbq}_{t})$ conserve energy well, the Hamiltonian $\BH(\hat{\mbp}_{t}, \hat{\mbq}_{t})$ should remain approximately constant over time. Consequently, the relative error in \Cref{eq:relerr-time} should remain close to zero.

\subsection{Non-Separable  Hamiltonian System}
\label{sec:nonseparable}
We compare H-FEX with SINDy \cite{SINDY} on a non-separable system from \cite{NonseparableProblem}.
The Hamiltonian system is characterized by the non-separable Hamiltonian function $\BH: \BR^2 \to \BR$ defined by: 
\begin{align}
\BH(p, q) = \exp(- \alpha _{1} p^{2}-\alpha_{2}q^{4}),
\label{eq:nonseparable-hamiltonian}
\end{align}
where $\alpha_{1}=1$ and $\alpha_{2}=1.1$.
By \Cref{eq:hamiltonian},
this Hamiltonian function yields the following dynamical system:
\begin{equation} \label{eq:nonseparable_dynamical}
    \begin{split}
        \frac{dp}{dt} &= 4\alpha _{2}q^{3} \exp(- \alpha _{1} p^{2}-\alpha_{2}q^{4}), \\ 
    \frac{dq}{dt} &= -2\alpha _{1}p \exp(- \alpha _{1} p^{2}-\alpha_{2}q^{4}).
    \end{split}
\end{equation} 
For the data, the initial states $(p_{t_0}, q_{t_0})$ are uniformly sampled from the domain $\Omega=[-1, 1]^2$. 
For the training data, we discretize the time interval $[0, 3]$ using $30$ points with a uniform timestep of $0.1$.
Using the adaptive step integrator RK45 \cite{RK45}, we generate 120 trajectories on the interval $[0, 3]$
from different initial states
using the true dynamical system \Cref{eq:nonseparable_dynamical}.
For the testing data, we use the same data generating process,
but instead use a longer time interval of $[0, 60]$,
discretized with the same timestep of $0.1$.

We state all the details regarding the structure and operator search loop of H-FEX for this non-separable system.
The H-FEX binary tree consists of three unary nodes and one binary node,
with 4 weights to scale the output of each node (see \Cref{fig:tree_structures}).
We use the following unary and binary operator dictionaries: 
    $$\unaryset =\{\mathrm{Id}, (\cdot)^2, (\cdot)^3,(\cdot)^4, \exp, \sin, (\cdot)^{-1}\},  \qquad
    \binaryset =\{+, \times,-, \div\},$$
where ``$\mathrm{Id}$" denotes the identity operator.
 We run the search loop for $100$ iterations, 
and each iteration,
we generate 15 FEX trees.
For operator sequence generation, 
we use an $\epsilon$-greedy strategy \cite{EpsilonGreedyStrategy} with $\epsilon=0.2$ to sample 15 operator sequences from the PMFs given by the controller.
During score computation,
we minimize the loss function \Cref{eq:empirical_loss}, for each FEX tree,
using the   Adam optimizer \cite{Adam} with a learning rate of $0.1$ for $150$ steps.
Since the system is non-separable,
we generate trajectories during training with \Cref{eq:train_integrator} using RK2 \cite{RK2},
a traditional integrator,
with $K=20$ substeps. 
The controller is a small fully-connected neural network deterministically mapping the zero vector to PMFs for each node,
and the weights of the controller are updated using the risk-seeking policy gradient \cite{RiskSeekingPolicyGradient} with $\nu=0.25$.
During the search loop, we maintain a candidate pool to save the $15$ top-scoring FEX models, 
and after the search loop, we fine-tune each of the models using Adam with a learning rate of $0.001$ for $300$ steps.
After the fine-tuning, we select the H-FEX tree with the highest score to compare with SINDy. 

\begin{figure}
    \centering
    \includegraphics[width=0.49\linewidth]{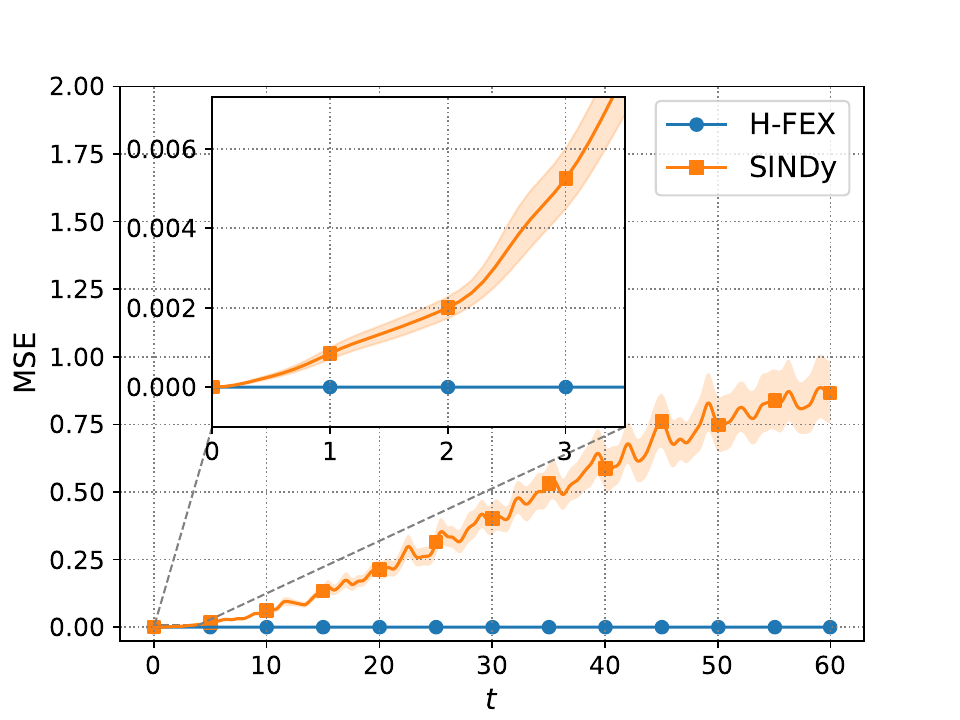}
    \includegraphics[width=0.49\linewidth]{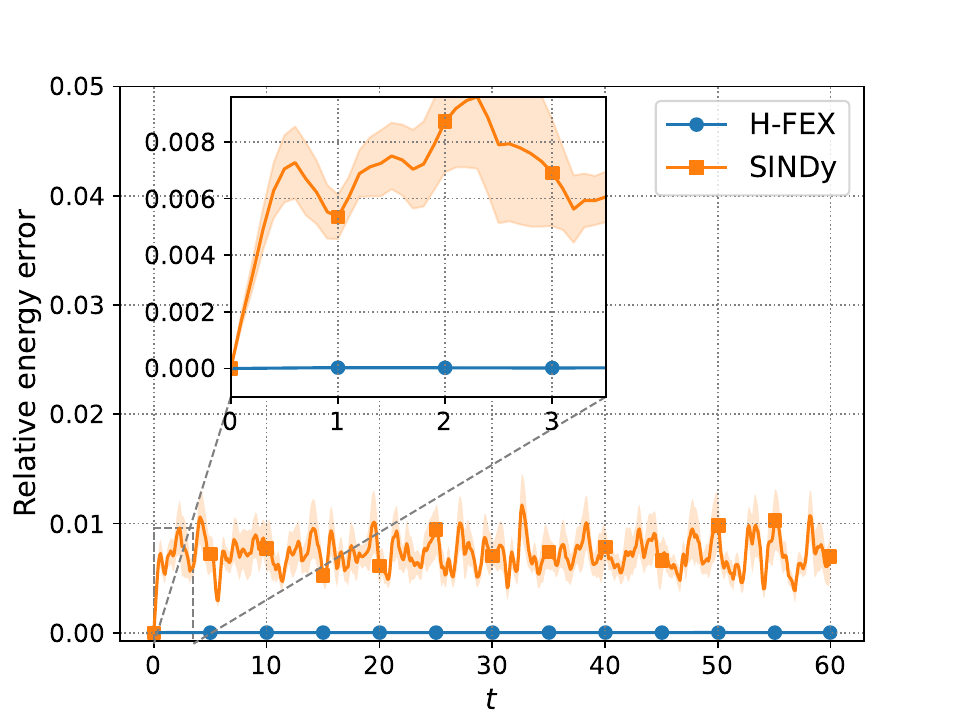}
    \caption{\textbf{Non-separable results.} FEX and SINDy are trained using trajectories on $[0, 3]$ and evaluated using $30$ test trajectories on $[0, 60]$. 
    \textbf{Left}: MSE over time, defined in \Cref{eq:MSE-time}.
    \textbf{Right}: Relative energy change over time, defined in \Cref{eq:relerr-time}. FEX exhibits virtually no MSE and energy drift over longer time horizons, in stark contrast to SINDy, which accumulates error over time and deviates from the initial energy almost immediately.}
    \label{fig:nonseparable-30pts}
\end{figure}

The resulting H-FEX tree has an operator sequence $\Be=((\cdot)^2, +, (\cdot)^4, \exp)$ (inorder traversal) and weights
$\Btheta=(0.9236, 1.0159, -1.083, 1.000)$,
yielding the closed form approximation of the Hamiltonian function:
\begin{align*}
\label{eq:nonseparable_hfex_eqn}
    \hat{\BH}_{\BTheta}(p, q) &=1.000 \exp(-1.083(0.9236 p^2 + 1.0159q^4)) \\
    &= \exp(-1.0002588 p^2 - 1.1002197 q^4).
\end{align*}
Comparing the approximation with \Cref{eq:nonseparable-hamiltonian}, H-FEX correctly identifies the true operators and learns weights that simplify to values close to the true values of $\alpha_1$ and $\alpha_2$.
Using \Cref{eq:predicted_trajectory} on the $30$ test trajectories, 
we generate predicted trajectories on the interval $[0, 60]$ given the initial states of the testing set.
We then use these predicted trajectories to compute the MSE and relative energy change over time and compare with SINDy in \Cref{fig:nonseparable-30pts}.
H-FEX maintains low trajectory error even at time points far beyond the training interval, whereas SINDY's trajectory error steadily increases over time.
Additionally, H-FEX preserves the initial energy throughout all time points, while SINDy quickly deviates from the initial energy.

On the other hand, SINDy yields a closed-form expression for $(\frac{dp}{dt}, \frac{dq}{dt})$ directly: 
\begin{align*}
    \frac{dp}{dt} &= 0.152 q - 0.683 p^2 q + 4.085 q^3 + 0.261 p^4 q - 0.834 p^2 q^3 - 2.752 q^5, \\ 
    \frac{dq}{dt} &= -1.901 p + 1.557 p^3 + 0.709 p q^2 - 0.365 p^5 - 0.593 p^3 q^2 + 0.457 p q^4.
\end{align*}
This learned expression differs significantly from the true dynamical system in \Cref{eq:nonseparable_dynamical}. The source of this discrepancy becomes clear upon brief examination of the SINDy algorithm \cite{SINDY}. 
SINDy constructs expressions by using a predefined library of candidate functions (e.g., polynomials, trigonometric terms) and selecting a sparse linear combination that best fits the observed data $(\mbp_{t_n}, \mbq_{t_n})_{n=0}^N$. 
In this experiment, the library includes polynomial terms up to degree $6$. 
Because SINDy can only produce linear combinations of these candidate functions, it struggles to capture complex nonlinear structures such as those present in \Cref{eq:nonseparable_dynamical}.
% The SINDy algorithm can only build linear combinations of candidate functions and therefore struggles with complex expressions like \cref{eq:nonseparable_dynamical}.

\subsection{Three-Body Problem}
\label{sec:three_body_problem}
The three-body problem \cite{ThreeBodyProblem} is a classical problem in physics and celestial mechanics that involves predicting the motion of three bodies interacting under their mutual gravitational forces.
In our experiments, we consider three point masses interacting in two-dimensional space.
For $1 \le i \le 3$, the $i$th body has its momentum denoted by $\mbp_i \in \BR^2$ and position denoted by $\mbq_i \in \BR^2$, 
so the system can be represented by coordinates $(\mbp_1,\mbp_2, \mbp_3,\mbq_1, \mbq_2, \mbq_3) \in \BR^{12}$.
The corresponding Hamiltonian function is $\BH: \BR^{12} \to \BR$ defined by:
\begin{equation}
    \BH(\mbp_1, \mbp_2, \mbp_3, \mbq_1, \mbq_2, \mbq_3)=
    \frac{\lVert \mbp_1 \rVert^2}{2m_1}+ \frac{\lVert\mbp_2 \rVert^2}{2m_2}+ \frac{\lVert\mbp_3 \rVert^2}{2m_3}    
    -\frac{Gm_1m_2}{\lVert \mbq_1 - \mbq_2 \rVert }-\frac{Gm_1m_3}{\lVert \mbq_1 - \mbq_3 \rVert }-\frac{Gm_2m_3}{\lVert \mbq_2 - \mbq_3 \rVert } 
    ,
    \label{eq:three_body_hamiltonian}
\end{equation}
where $\norm{\cdot}$ denotes the Euclidean norm,
$m_i \in \BR$ denotes the mass of the $i$th body,
and $G \in \BR$ denotes the gravitational constant. 
In our experiments, we set $m_i=1$ for each $i$ and $G=1$.
To generate initial states,
we place the three point masses on a randomly chosen circle (with radius uniformly drawn from $[0.9, 1.2]$)
by selecting a random point for the first point mass and then rotating it by $120^{\circ}$ for the other two.
Each point mass is given a momumtum perpendicular to its position vector to produce a perfect circular orbit. 
Finally, we perturb these momenta by multiplying them by a random factor uniformly drawn from $[0.8, 1.2]$.
For testing, we generate $30$ trajectories on the time interval $[0, 30]$ with a timestep of $0.1$ using the integrator RK45 \cite{RK45} until a relative error less than $10^{-9}$ is attained.
%However, for training, we consider two different datasets.

\begin{figure}[H]
    \centering
    \includegraphics[width=0.75\linewidth]{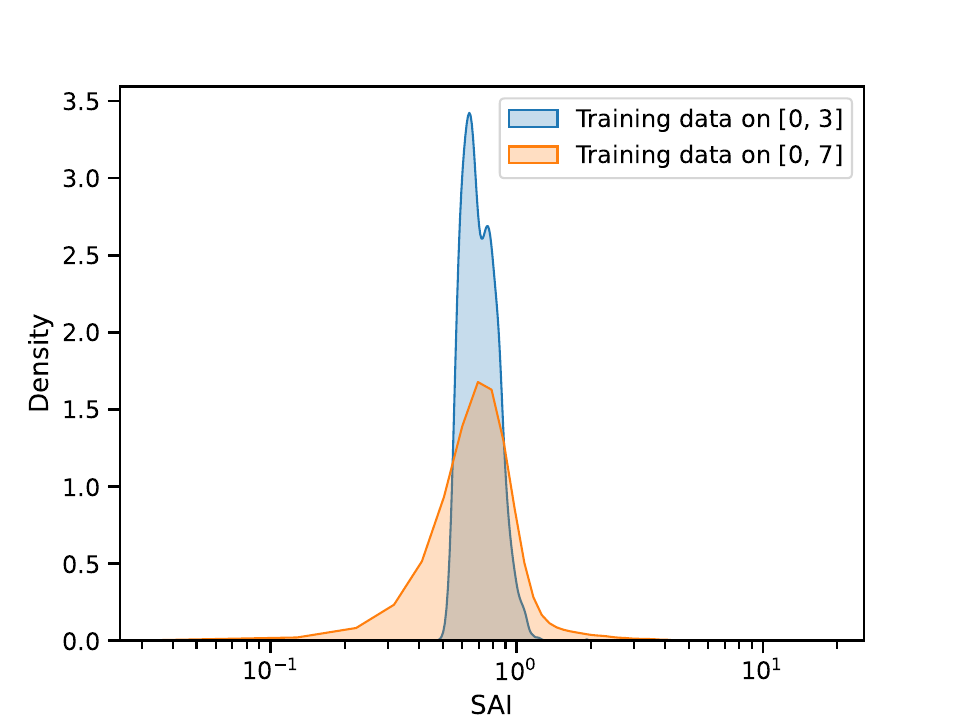}
    \caption{\textbf{SAI Density Plot.} The stiffness-aware index (SAI), defined in \Cref{eq:SAI}, is used to empirically quantify the stiffness of training trajectories under two different scenarios in the three-body problem. The plot compares trajectories on the intervals $[0, 3]$ and $[0, 7]$.
    Higher SAI values correspond to stiffer regions of the trajectory. As shown in the density plots, trajectories on $[0, 7]$ exhibit stiff regions not present in $[0, 3]$, primarily due to collisions occurring over the longer interval.}
    \label{fig:body3-SAI}
\end{figure}

We consider two different training datasets:
the first with trajectories on a smaller time interval $[0, 3]$,
and the second with trajectories on a larger time interval $[0, 7]$.
As the trajectories are being generated, collisions between two or more point masses will inevitibly occur, with more potential for collisions on larger time intervals.
Therefore, trajectories on the interval $[0, 3]$ tend to have less collisions than trajectories on the interval $[0, 7]$.
When collisions occur, two or more position vectors become very close, leading to a singularity in \Cref{eq:three_body_hamiltonian}.
In other words, collisions create stiff regions in a trajectory, where integration becomes more challenging.
To quantify the stiffness of the two training datasets, we calculate the stiffness-aware index (SAI) from \cite{SANN,EmpiricalStiffness} defined by:
\begin{equation}
    \mathrm{SAI}(\mbp_{t_i}, \mbq_{t_i}) \coloneqq \frac{1}{||(\mbp_{t_i}, \mbq_{t_i})^\top||} \cdot \frac{||(\mbp_{t_{i+1}}, \mbq_{t_{i+1}})^\top - (\mbp_{t_i}, \mbq_{t_i})^\top||}{t_{i+1} - t_i},
    \label{eq:SAI}
\end{equation}
where $\norm{\cdot}$ is the Euclidean norm and $(\mbp_{t_{i}},   \mbq_{t_{i}})^\top \in \BR^{2d}$ is the state of the trajectory at time $t_i$.
Higher SAI values correspond to trajectory segments that exhibit greater stiffness.
In \Cref{fig:body3-SAI}, we show the densities of both datasets.
The dataset on $[0, 7]$ has regions with larger SAI compared to the dataset on $[0, 3]$, 
supporting our intuition that trajectories on the interval $[0, 7]$ are more challenging to integrate due to increased presence of collisions. 
For both training datasets, 
we use the same H-FEX structure and training hyperparameters.
% {\color{redf} what does $\|\cdot\|$ mean and how to calculate? what does $(\cdot)^T$ mean?}

The structure of an H-FEX tree consists of three unary nodes, one binary node, and one interaction node to produce interaction terms among the coordinate terms $(\mbq_1, \mbq_2, \mbq_3)$,
with weights after each binary and unary node (see \Cref{fig:tree_structures}).
For the unary, binary, and interaction nodes, we use the following operator dictionaries:
\begin{align*}
    \unaryset&=\{ \mathrm{Id}, (\cdot)^2, (\cdot)^3, \exp, \sin, (\cdot)^{-1} \}, \\
    \binaryset&=\{ +, \times, - \}, \\
    \interactionset &=\{ \norm{\mbq_i - \mbq_j}^2, \norm{\mbq_i - \mbq_j}, \norm{\mbq_i \odot \mbq_j}^2, \norm{\mbq_i \odot \mbq_j} \} \quad \text{for $i \ne j$}. \\
\end{align*}  
We run the search loop for $100$ iterations, and for each iteration, we sample $10$ operator sequences from the controller using an $\epsilon$-strategy \cite{EpsilonGreedyStrategy} with $\epsilon=0.2$. 
We compute the score by minimizing \Cref{eq:empirical_loss} using Adam \cite{Adam} with a learning rate of $0.1$ for $150$ steps.
Predicted trajectories are generated during training using \Cref{eq:train_integrator} with $K=20$ substeps using leapfrog \cite{Leapfrog}, a symplectic integrator.
The weights of the controller are updated using the risk-seeking policy gradient in \Cref{eq:risk-seeking-gradient} with $\nu=0.25$.
The candidate pool saves the top $15$ highest-scoring H-FEX trees,
and after the search loop, we fine-tune each of the trees using Adam with a learning rate of $0.001$ for $300$ steps.
After fine-tuning, we select the highest scoring H-FEX tree to compare with other models.

We show only the results of H-FEX when trained with trajectories on $[0, 7]$,
as the results of H-FEX when trained with trajectories on $[0, 3]$ are similar for H-FEX.
The highest-scoring H-FEX tree has an operator sequence of $\Be = \left( (\cdot)^2, +, \norm{\mbq_i - \mbq_j}, (\cdot)^{-1}, \mathrm{Id} \right)$, where $(\cdot)^2, (\cdot)^{-1}, \mathrm{Id} \in \unaryset$, $+\in \binaryset$, and $\norm{\mbq_i - \mbq_j} \in \interactionset$.
This yields the closed form approximation of the Hamiltonian function:
\begin{align*}
    \hat{\BH}_{\BTheta}(\mbp_1, \mbp_2, \mbp_3, \mbq_1, \mbq_2, \mbq_3) = 
    & \left( 
    0.5005 p_{1,x}^2 + 0.4981 p_{1, y}^2 
    + 0.4996 p_{2, x}^2 + 0.4985 p_{2, y}^2
    + 0.4976 p_{3, x}^2 + 0.5024 p_{3, y}^2
    \right)
    \\ 
    & +
    \left(
    \frac{-0.9985}{\norm{\mbq_1 - \mbq_2}} + \frac{-0.9970}{\norm{\mbq_1 - \mbq_3}} + \frac{-0.9949}{\norm{\mbq_2 - \mbq_3}} 
    \right),
\end{align*}
where for each body, $\mbp_i = (p_{i, x}, p_{i, y})^\top$ is the momentum on the x-y plane. 
The equation learned by H-FEX closely matches \Cref{eq:three_body_hamiltonian},
as we let $G=1$ and $m_i=1$ for each body in our experiments.
With \Cref{eq:predicted_trajectory},
we use $30$ intial conditions from the test dataset to generate
predicted trajectories on the interval $[0, 30]$
using the highest-scoring H-FEX tree
and compare to other models for both training scenarios. 

\begin{figure}[H]
    \centering
    \includegraphics[width=0.49\linewidth]{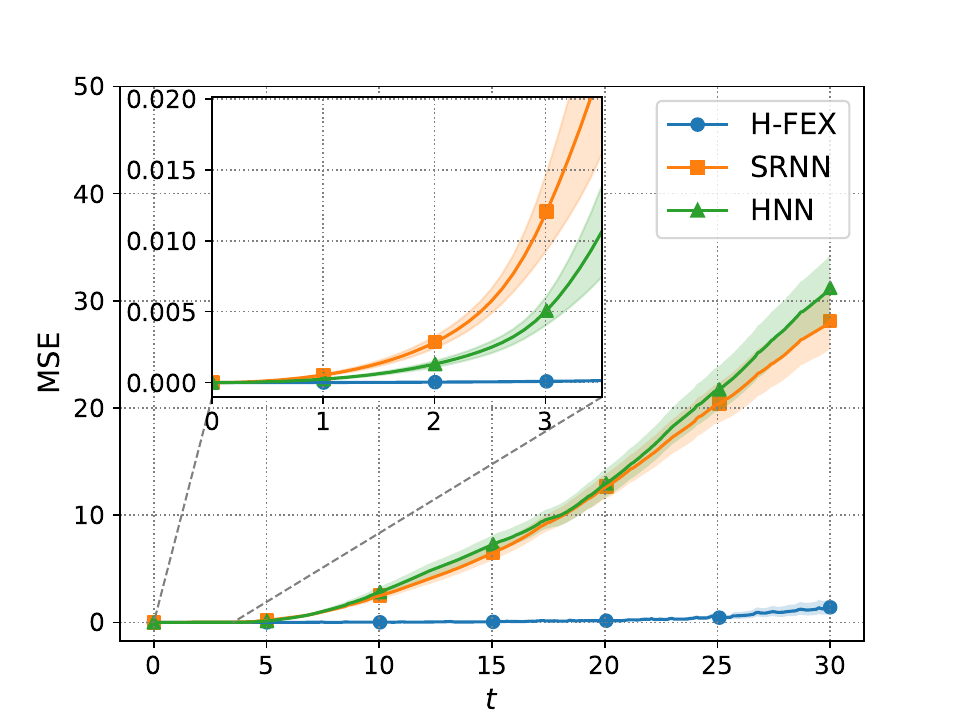}
    \includegraphics[width=0.49\linewidth]{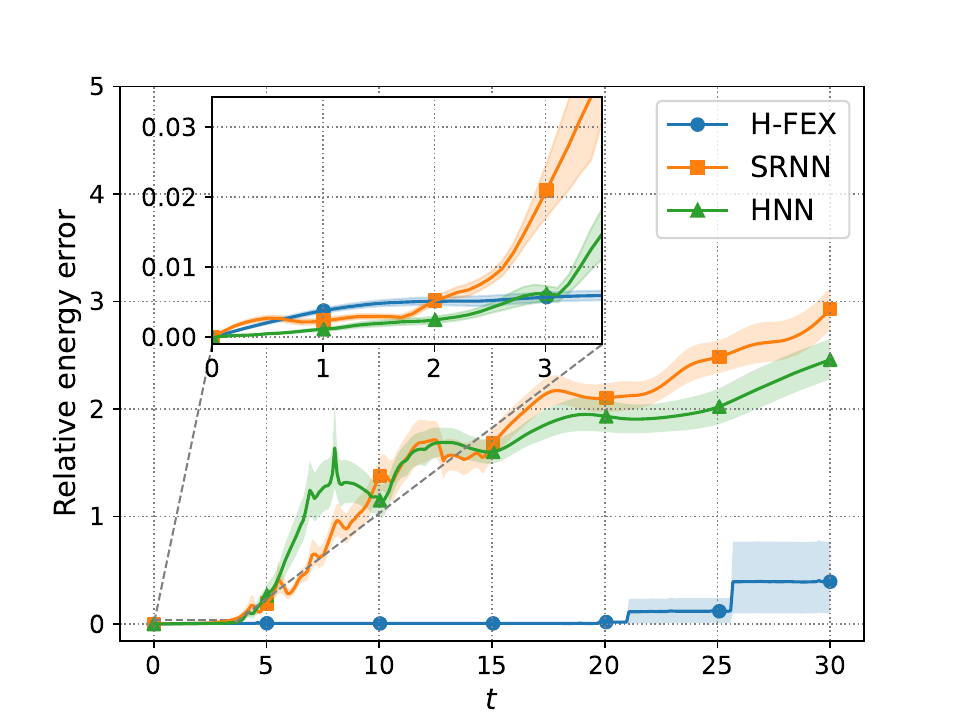}
    \caption{\textbf{Three-body problem results with training on $[0, 3]$.}
    H-FEX, SRNN, and HNN are trained using trajectories on $[0, 3]$ and evaluated using $30$ test trajectories on $[0, 30]$. 
    \textbf{Left}:  MSE over time, defined in \Cref{eq:MSE-time}. 
    \textbf{Right}: Relative energy change over time, defined in \Cref{eq:relerr-time}.
    H-FEX shows slower growth in both MSE and enery drift over time compared to SRNN and HNN. } 
    \label{fig:body3-30pts}
\end{figure}

In \Cref{fig:body3-30pts}, we plot the MSE and relative energy change over time for models trained with trajectories on the interval $[0, 3]$.
We compare H-FEX with SRNN \cite{SRNN} and HNN \cite{HNN}, which are both methods that use a neural network as a surrogate for the Hamiltonian function so an interpretable closed-form expression is not attainable.
H-FEX maintains accurate trajectories even at large time points,
and it preserves the initial energy for long periods of time
with deviations in the energy only occuring at later time points.

\begin{figure}[H]
    \centering
    \includegraphics[width=0.49\linewidth]{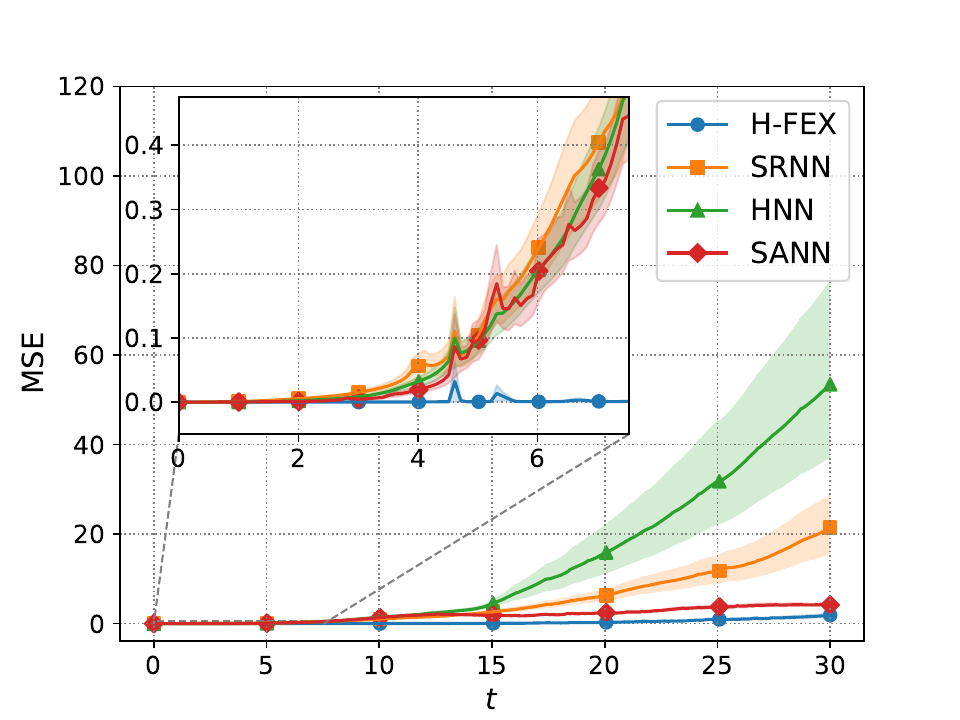}
    \includegraphics[width=0.49\linewidth]{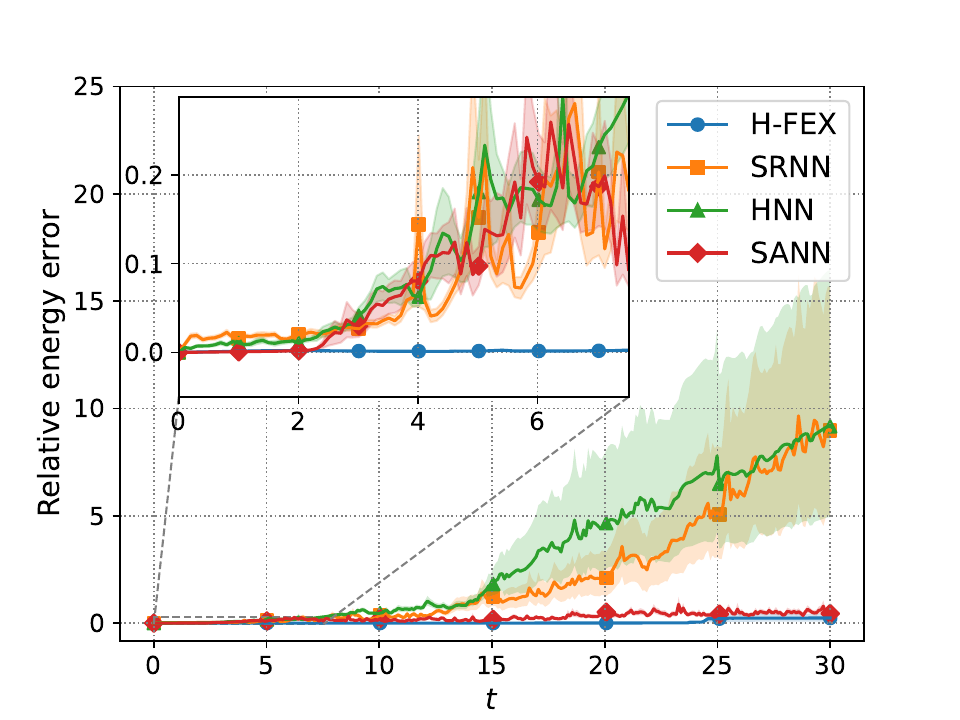}
    \caption{\textbf{Three-body problem results with training on [0, 7].} 
    H-FEX, SRNN, HNN, and SANN are trained using trajectories on $[0, 7]$, which are more stiff due to collisions, and evaluated using $30$ test trajectories on $[0, 30]$.
    \textbf{Left}:  MSE over time, defined in \Cref{eq:MSE-time}.
    \textbf{Right}: Relative energy change over time, defined in \Cref{eq:relerr-time}.
     Despite the increased difficult of the training data, H-FEX maintains the lowest MSE and energy drift among all methods.} 
    \label{fig:body3-70pts} 
\end{figure}

In \Cref{fig:body3-70pts}, we plot the MSE and relative energy change over time for models trained with trajectories on the interval $[0, 7]$.
Here, we compare H-FEX with SRNN, HNN, and SANN \cite{SANN},
where SANN is a neural network surrogate for the Hamiltonian function
that can more accurately learn stiff regions of the training trajectories.
Since the dataset on $[0, 7]$ contains more collisions, the
MSE and the relative energy change of HNN and SRNN diverge faster
than when trained using the dataset on $[0, 3]$.
H-FEX still produces the most accurate trajectories and preserves energy the best,
while providing a closed-form approximation of the Hamiltonian function. 

\section{Conclusion} \label{sec:conclusion}
This paper introduces H-FEX, an adaptation of FEX for Hamiltonian systems.
H-FEX represents the Hamiltonian function as a tree of operators, enabling it to learn complex mathematical expressions.
We modify the FEX search loop for Hamiltonian systems
and introduce interaction nodes to better capture interaction terms common in coupled systems.
Numerical experiments show that H-FEX accurately identifies the operators and weights needed to construct a tree that closely approximates the true Hamiltonian function.
Moreover, the predicted trajectories are highly accurate and preserve energy over long time horizons, 
demonstrating H-FEX’s capability in accurately modeling complex Hamiltonian systems.

Like symbolic regression methods such as SINDy, H-FEX depends on the choice of operators in its dictionaries, which significantly impacts the accuracy of the learned closed-form solution. If essential operators are missing, H-FEX’s accuracy will suffer. Additionally, H-FEX requires a predefined tree structure with operator nodes and weights, shaping the resulting equations. When prior knowledge of the Hamiltonian system is available, it can guide the selection of operators and tree structure to improve modeling. However, without prior knowledge, one must experiment with various operators and structures. In future work, we believe it is fruitful to develop adaptive methods that dynamically adjust node and weight placement, allowing H-FEX to learn complex physics without a predefined tree structure.

\section*{Acknowledgement}
The research of Senwei Liang was partially supported by the U.S. Department of Energy, Office of Science, Office of Advanced Scientific Computing Research SciDAC program under Contract No. DE-AC02-05CH11231.  The research of Chunmei Wang was partially supported by National Science Foundation Grant DMS-2206332.

\bibliographystyle{unsrt} 
\bibliography{references}

\newpage
\appendix

\section{Additional Integrator Details} \label{apx:integrator_details}

This section provides further information on the integrators used during both evaluation (see \Cref{eq:eval_integrator}) and training (see \Cref{eq:train_integrator}).
To reduce notational clutter, we use the shorthand $(\mbp_n, \mbq_n)$ to denote $(\mbp_{t_n}, \mbq_{t_n})$.
The notation $\hat{\BH}_p(\mbp_k, \mbq_k)$ denotes the derivative of the surrogate Hamiltonian $\hat{\BH}$ with respect to $\mbp$, evaluated at the point $(\mbp_k, \mbq_k)$.
This derivative can be computed exactly and quickly using automatic differentiation.

The multi-step integration algorithms for advancing a single time step using the Leapfrog and RK2 integrators are shown in \Cref{alg:leapfrog} and \Cref{alg:rk2} respectively. 
These algorithms take the surrogate of the Hamiltonian function $\hat{\BH}$ and the current state $(\tilde{\mbp}_n, \tilde{\mbq}_n)$, integrates for $K$ substeps, and outputs the predicted next state $(\hat{\mbp}_{n+1}, \hat{\mbq}_{n+1})$.
During evaluation, the current state $(\tilde{\mbp}_n, \tilde{\mbq}_n)$ is set to the previous predicted state $(\hat{\mbp}_{n}, \hat{\mbq}_{n})$,
while during training, the current state $(\tilde{\mbp}_n, \tilde{\mbq}_n)$ is set to the previous observed state $(\mbp_{n}, \mbq_{n})$.
It should be noted that the integration for each time step is parallelizable during training, greatly reducing the runtime for computing a score of an H-FEX tree.

The Leapfrog and RK2 integrators have different properties and assumptions.
The Leapfrog integrator belongs to the class of symplectic integrators, which simulate trajectories with less energy drift, but assumes the Hamiltonian function is separable, i.e., $\BH(\mbp,\mbq)=K(\mbp)+U(\mbq)$. 
On the other hand, the RK2 integrator does not rely on the separability of the Hamiltonian but lacks the energy drift reduction properties of Leapfrog integration. 
Consequently, we first attempt to use Leapfrog during training, but if many H-FEX trees have trouble optimizing, we switch to RK2 as a fallback.

\begin{algorithm}
    \caption{Multi-step Leapfrog integrator for Hamiltonian systems}
    \label{alg:leapfrog}
    \begin{algorithmic}
        \Function{LeapfrogIntegrator}{$\hat{\BH}$, $(\tilde{\mbp}_n, \tilde{\mbq}_n)$, $(t_{n_k})_{k=1}^K$}
            \State $\Delta t \gets t_{n_2} - t_{n_1}$ \Comment{$(t_{n_k})_{k=1}^K$ are uniform substeps}
            \State $(\mbp_0, \mbq_0)\gets(\tilde{\mbp}_n, \tilde{\mbq}_n)$
            \For{$k=0$ to $K-1$} 
                \State $\mbp_{k+\frac{1}{2}} \gets \mbp_k - \frac{\Delta t}{2} \hat{\BH}_q(\mbp_k, \mbq_k)$ \Comment{Leapfrog update adapted for Hamiltonian systems}
                \State $\mbq_{k+1} \gets \mbq_k + \Delta t \hat{\BH}_p(\mbp_{k+\frac{1}{2}}, \mbq_k)$
                \State $\mbp_{k+1} \gets \mbp_{k+\frac{1}{2}} - \frac{\Delta t}{2} \hat{\BH}_q(\mbp_{k+\frac{1}{2}}, \mbq_{k+1})$
            \EndFor
            \State $(\hat{\mbp}_{n+1}, \hat{\mbq}_{n+1})\gets(\mbp_K, \mbq_K)$
            \State \Return $(\hat{\mbp}_{n+1}, \hat{\mbq}_{n+1})$
        \EndFunction
    \end{algorithmic}
\end{algorithm}

\begin{algorithm}
    \caption{Multi-step RK2 integrator for Hamiltonian systems}
    \label{alg:rk2}
    \begin{algorithmic}
        \Function{RK2Integrator}{$\hat{\BH}$, $(\tilde{\mbp}_n, \tilde{\mbq}_n)$, $(t_{n_k})_{k=1}^K$}
            \State $\Delta t \gets t_{n_2} - t_{n_1}$ \Comment{$(t_{n_k})_{k=1}^K$ are uniform substeps}
            \State $(\mbp_0, \mbq_0)\gets(\tilde{\mbp}_n, \tilde{\mbq}_n)$
            \For{$k=0$ to $K-1$} 
                \State $r_1^{(p)} \gets -\hat{\BH}_q(\mbp_k, \mbq_k)$ \Comment{RK2 update adapted for Hamiltonian systems}
                \State $r_1^{(q)} \gets \hat{\BH}_p(\mbp_k, \mbq_k)$
                \State $r_2^{(p)} \gets -\hat{\BH}_q(\mbp_k + \frac{\Delta t}{2} r_1^{(p)}, \mbq_k + \frac{\Delta t}{2} r_1^{(q)})$
                \State $r_2^{(q)} \gets \hat{\BH}_p(\mbp_k + \frac{\Delta t}{2} r_1^{(p)}, \mbq_k + \frac{\Delta t}{2} r_1^{(q)})$
                \State $\mbp_{k+1} \gets \mbp_k + \Delta t \, r_2^{(p)}$
                \State $\mbq_{k+1} \gets \mbq_k + \Delta t \, r_2^{(q)}$
            \EndFor
            \State $(\hat{\mbp}_{n+1}, \hat{\mbq}_{n+1})\gets(\mbp_K, \mbq_K)$
            \State \Return $(\hat{\mbp}_{n+1}, \hat{\mbq}_{n+1})$
        \EndFunction
    \end{algorithmic}
\end{algorithm}

\section{H-FEX Search Loop}

In \Cref{alg:fex_loop}, we show a detailed pseudocode of the H-FEX search loop. 
The search loop seeks the optimal operator sequence in $E$ iterations. 
Each iteration, PMFs are generated by the controller $\chi_\Phi$, and $B$ H-FEX trees are constructed from sampled operator sequences.
After constructing an H-FEX tree, the weights are optimized for $L$ iterations. 
The optimization of $B$ H-FEX trees is independent and can be easily parallelized.
The score of each of the $B$ operator sequences can then be used to compute the risk seeking policy gradient in \Cref{eq:risk-seeking-gradient}.

\begin{algorithm}
    \caption{H-FEX search loop high-level pseudocode}
    \label{alg:fex_loop}
    \begin{algorithmic}
    \Procedure{SearchLoop}{$\tree$} \Comment{Assume a fixed tree $\mathcal{T}$}
        \For{$1$ to $E$} 
            \State $\mathrm{PMFs} \gets \chi_\Phi(\mathbf{0})$
            \For{$1$ to $B$} \Comment{Constructing multiple trees with sampled operator sequences}
                \State $\Be \sim \mathrm{PMFs}$ 
                \State $\Btheta \sim \mathrm{UniformDistribution}$ \Comment{Initializing network weights}
                \State Construct H-FEX tree $\hat{\BH}(\cdot, \cdot;\tree,\Be,\Btheta)$
                \For{$1$ to $L$} \Comment{Optimizing weights}
                    \State $(\hat{\mbp}, \hat{\mbq})\gets \Integrator(\hat{\mathcal{H}}, (\mbp_0, \mbq_0), )$
                    \State Compute loss $\mathcal{L}(\BTheta)$
                    \State Update tree weights $\Btheta$
                \EndFor
                \State Compute score $S(\Be)$
                \If{$S(\Be)$ among top $K$ scoring models in the pool}
                    \State Add $\hat{\BH}$ to the pool
                \EndIf
            \EndFor
            \State Compute risk-seeking policy gradient $\nabla_\Phi \mathcal{J}(\Phi)$
            \State Update controller weights $\Phi$
        \EndFor
        \EndProcedure
    \end{algorithmic}
\end{algorithm}

\section{H-FEX Tree Structures}
\begin{figure}[H]
    \centering
    \includegraphics[width=0.7\linewidth]{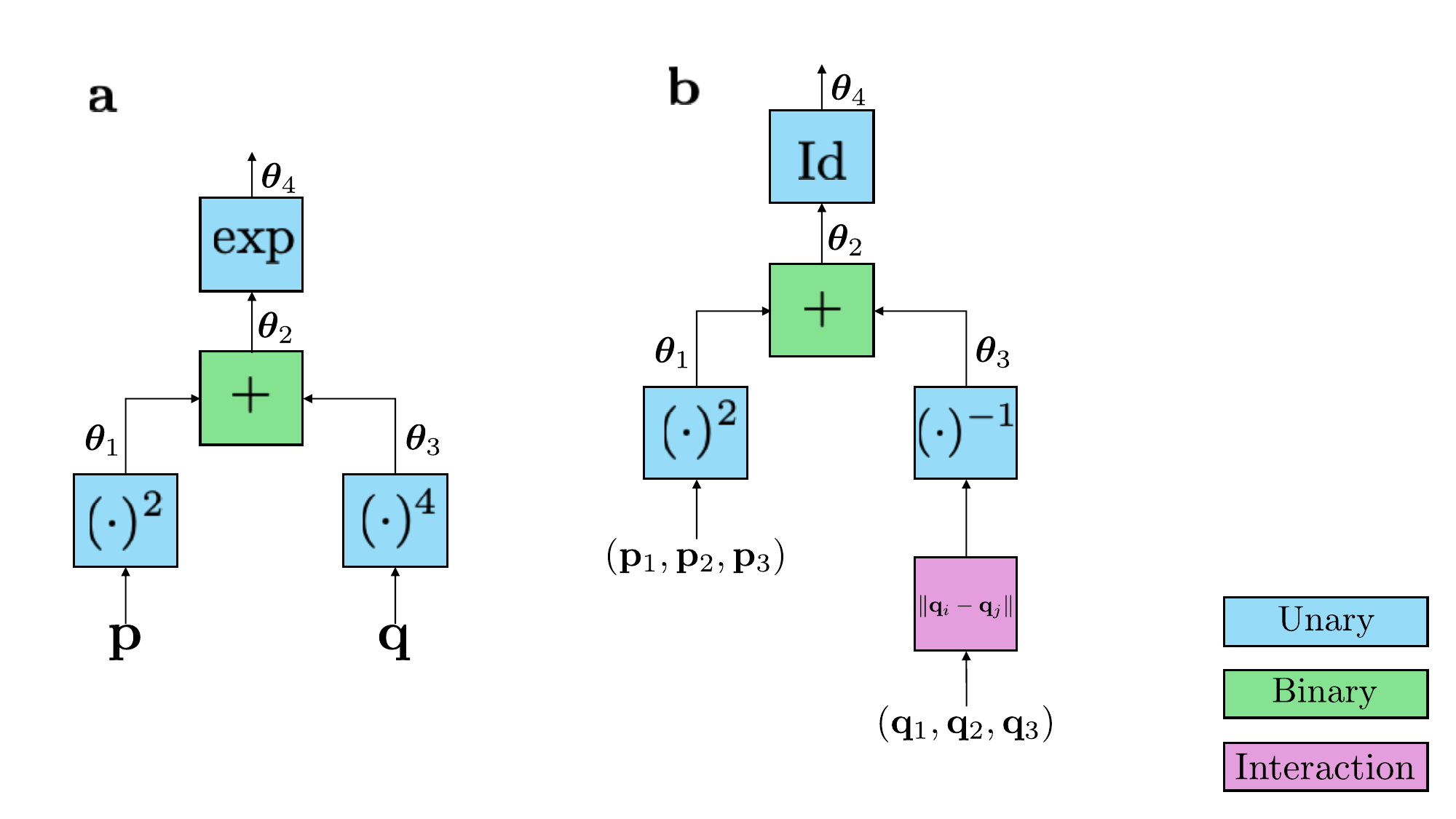}
    \caption{\textbf{Tree structure of H-FEX for numerical experiments}. \textbf{(a)} shows the tree structure and final learned operator sequence of H-FEX for the nonseparable experiment in \Cref{sec:nonseparable}. \textbf{(b)} shows the tree and final learned operators.}
    \label{fig:tree_structures}
\end{figure}

We show the H-FEX tree structures in \Cref{fig:tree_structures} used for each of the numerical experiments in \Cref{sec:numerical_results}.
The tree structure is fixed before training, and the operators are the final learned operators.
Weights are added after each of the unary and binary nodes.

\end{document}